\documentclass[lettersize,journal]{IEEEtran}
\usepackage{amsmath,amsfonts}
\usepackage{array}
\usepackage[caption=false,font=normalsize,labelfont=sf,textfont=sf]{subfig}
\usepackage{textcomp}
\usepackage{stfloats}
\usepackage{multirow}
\usepackage{url}
\usepackage{verbatim}
\usepackage{graphicx}
\usepackage{booktabs}
\usepackage{xcolor}
\usepackage{cite}
\usepackage{amsthm}
\usepackage{algorithm}
\usepackage{algpseudocode}

\begin{document}

\title{A PPO-Based Bitrate Allocation Conditional Diffusion Model for Remote Sensing Image Compression}

\author{Yuming~Han,~\IEEEmembership{Graduate Student Member,~IEEE,}
        Jooho~Kim, Anish~Shakya,~\IEEEmembership{Graduate Student Member,~IEEE}
\thanks{Yuming Han is with the Department of Electrical and Computer Engineering, Texas A$\&$M University, College Station, TX 77843 USA (e-mail: yuminghan@tamu.edu).}
\thanks{Jooho Kim is with the Institute for a Disaster Resilient Texas, Texas A$\&$M University, College Station, TX 77843 USA (e-mail: Jooho.kim@tamu.edu).}
\thanks{Anish Shakya is with Department of Marine $\&$ Coastal Environmental Science, Texas A$\&$M University, Galveston, TX 77554 USA (anish$\_$shakya@tamu.edu).}
}



\maketitle

\begin{abstract}
Existing remote sensing image compression methods still explore to balance high compression efficiency with the preservation of fine details and task-relevant information. Meanwhile, high-resolution drone imagery offers valuable structural details for urban monitoring and disaster assessment, but large-area datasets can easily reach hundreds of gigabytes, creating significant challenges for storage and long-term management. In this paper, we propose a PPO-based bitrate allocation Conditional Diffusion Compression (PCDC) framework. PCDC integrates a conditional diffusion decoder with a PPO-based block-wise bitrate allocation strategy to achieve high compression ratios while maintaining strong perceptual performance. We also release a high-resolution drone image dataset with richer structural details at a consistent low altitude over residential neighborhoods in coastal urban areas. Experimental results show compression ratios of 19.3$\times$ on DIV2K and 21.2$\times$ on the drone image dataset. Moreover, downstream object detection experiments demonstrate that the reconstructed images preserve task-relevant information with negligible performance loss.
\end{abstract}


\section{Introduction}
With the ongoing advances in remote sensing (RS) technology, the volume of imagery collected by modern aerial platforms, especially unmanned aerial vehicles (UAVs), is increasing dramatically~\cite{hu2019uav}. This growth is fueled by the rapid adoption of high resolution sensors and more frequent data acquisition, which together produce massive datasets with rich spatial details~\cite{intro_1}. 
Efficient compression is therefore essential for reducing storage overhead and easing the burden of data management, while also enabling more scalable post collection processing~\cite{intro_2}. 
% In practical pipelines, the challenge is not only achieving high compression efficiency, but also keeping the offline encoding and decoding cost manageable, since large archives can make compression time and computational consumption meaningful constraints in the overall workflow~\cite{intro_3}. 
% Moreover, the decompressed imagery should preserve task relevant information as faithfully as possible, ideally maintaining or even improving performance for downstream analytics such as object detection, so that storage savings do not come at the expense of operational utility.
Traditional image compression methods, such as JPEG2000~\cite{taubman2002jpeg2000} and BPG~\cite{BellardBPG2018}, have been widely adopted in remote sensing workflows. 
% Most of these methods follow a broadly similar hand crafted pipeline that includes transform processing, quantization, and entropy coding. In brief, the input image is first mapped into a transform domain to reduce spatial correlation, then the transform coefficients are quantized to control the bitrate while retaining as much useful information as possible, and finally entropy coding is applied to further compress the resulting symbols. 
% Beyond these standards, predictive coding methods such as JPEG-LS~\cite{weinberger2000loco}, reduce redundancy by encoding prediction residuals. Vector quantization approaches use codebook based block representation and learning vector quantization schemes to map image blocks into compact indices for storage and transmission~\cite{wang2013vectorquantization}.
However, applying traditional codecs to drone based remote sensing imagery has three key limitations. First, remote sensing images contain dense textures, sharp edges, and strong scale variation, making them difficult to compress effectively with fixed transforms and uniform quantization~\cite{xing2023flood}. Second, traditional compression is mainly optimized for pixel level reconstruction quality~\cite{BellardBPG2018}, which does not preserve the most task relevant information. Third, at low bitrates, traditional codecs introduce visible distortions and lose fine structures that are critical for remote sensing analysis~\cite{pan2023hybrid}.

To overcome the limitations of traditional methods, recent studies have explored neural network based compression, such as convolutional neural network (CNN)~\cite{ACNN}, Transformer~\cite{Vit}, and generative adversarial network (GAN)~\cite{hific} architectures. 
% Most learned frameworks are built on autoencoder (AE) and variational autoencoder (VAE) style encoder-decoder architectures~\cite{AE_1,AE_2,VAE}, which introduces a probabilistic latent space and an entropy model to better support rate control and reconstruction. In practice, the encoder transforms the image into quantized latents that are entropy coded into a bitstream, and the decoder reconstructs the image from this compact representation. Building on the AE and VAE framework, researchers have strengthened the encoder and decoder with more powerful feature modeling backbones, leading to learned compression methods dominated by convolutional neural network (CNN), Transformer, and generative adversarial network (GAN) architectures. 
% Tang et al.~\cite{ACNN} incorporated graph attention with asymmetric convolutions to enhance global interaction beyond local receptive fields. Li et al.~\cite{Vit} developed a Vision Transformer (ViT) that tokenizes images into patches and uses Transformer blocks for both analysis and synthesis, demonstrating improved rate distortion performance at low bitrates. Han et al.~\cite{han2023edge} proposed an edge guided adversarial framework that constrains edge fidelity to preserve sharp structures and textures. 
However, these approaches still suffer from three limitations in remote sensing image compression: (i) deterministic single pass decoding limits error correction and causes over smoothing~\cite{ACNN}; (ii) fixed rate distortion objectives impose a rigid tradeoff between fidelity and perceptual quality~\cite{ACNN,Vit}; and (iii) reconstruction quality degrades rapidly at very low bitrates, leading to detail loss or unstable textures~\cite{hific}. To address these limitations, diffusion models (DMs)~\cite{ho2020denoising} have been introduced into learned image compression as a new class of generative decoders. Diffusion based reconstruction proceeds via iterative denoising, which provides a natural mechanism to correct quantization induced errors and recover missing structures~\cite{cdc}. Moreover, the sampling process enables more flexible control of the fidelity realism tradeoff, helping reduce over smoothing and suppress unnatural artifacts~\cite{ghouse2023residual}. Since these models operate in pixel space, the cost of inference is high due to sequential evaluation. The latent diffusion model (LDM)~\cite{rombach2022high} was introduced to reduce computational costs by performing diffusion and reverse steps in the latent space. 
% In~\cite{li2025exploring}, latent diffusion models were introduced to improve RS image compression by enhancing prior modeling and fine detail restoration. 
However, the decoding complexity and computational cost remain high for large scale RS datasets.

Beyond technical performance, the effect of compression on downstream tasks is also critical in remote sensing, since decompressed images are used for applications such as object detection and scene understanding~\cite{object_1}. However, most existing studies focus mainly on rate distortion or perceptual quality, while downstream task performance remains underexplored. 
% Notably, diffusion based decoders can better support such tasks by suppressing compression artifacts, enhancing object boundaries, and restoring more coherent textures. 
Therefore, to address the problems above, we present a diffusion model based and downstream task friendly drone image compression model. 
% Specifically, we build on the standard context dependent diffusion compression pipeline and explicitly capture the information lost during context reconstruction by computing the residual between the original image and the context encoder reconstruction, then compressing this residual in a sparse and quantized form with minimal overhead. During decompression, the recovered residual is added back to the diffusion reconstruction to restore high frequency details that are particularly important in UAV imagery, such as small objects, thin structures, and dense textures. Moreover, beyond rate distortion and perceptual evaluation, we further conduct object detection on the decompressed images to demonstrate that the proposed method better preserves task relevant cues and yields improved downstream performance.
The primary contributions of this article can be summarized
as follows.
\begin{itemize}
  \item \textbf{Conditional Diffusion Compression Framework:} We propose a PPO-based bitrate allocation Conditional Diffusion Compression (PCDC) framework that  considers both reconstruction fidelity and downstream task utility. Numerical results show that we achieve 19.3$\times$ compression on DIV2K and 21.2$\times$ on the drone image dataset.

  \item \textbf{PPO-based Bitrate Allocation:} We design a novel PPO--Lagrangian controller performs block-wise bitrate allocation using high-frequency residual and CNN encoder features as state inputs. The policy selects allocation levels for each block under a global bitrate constraint, achieving superior perceptual performance while maintaining the target compression ratio.

  \item \textbf{Object Detection on Self-collected Dataset:} We release a drone image dataset consisting of high-resolution nadir images captured at a low altitude with richer structural details in coastal urban areas. Furthermore, downstream object detection results show negligible differences between the original and reconstructed images, demonstrating that the PCDC effectively preserves task-relevant information.
\end{itemize}

The rest of this article is organized as follows. Section~\ref{sec:model} presents the system model, Section~\ref{sec:results} reports the experimental results, and Section~\ref{sec:conclusion} concludes the article.

\section{System Model}
\begin{figure*}[bth]
    \centering
    \includegraphics[width=\linewidth]{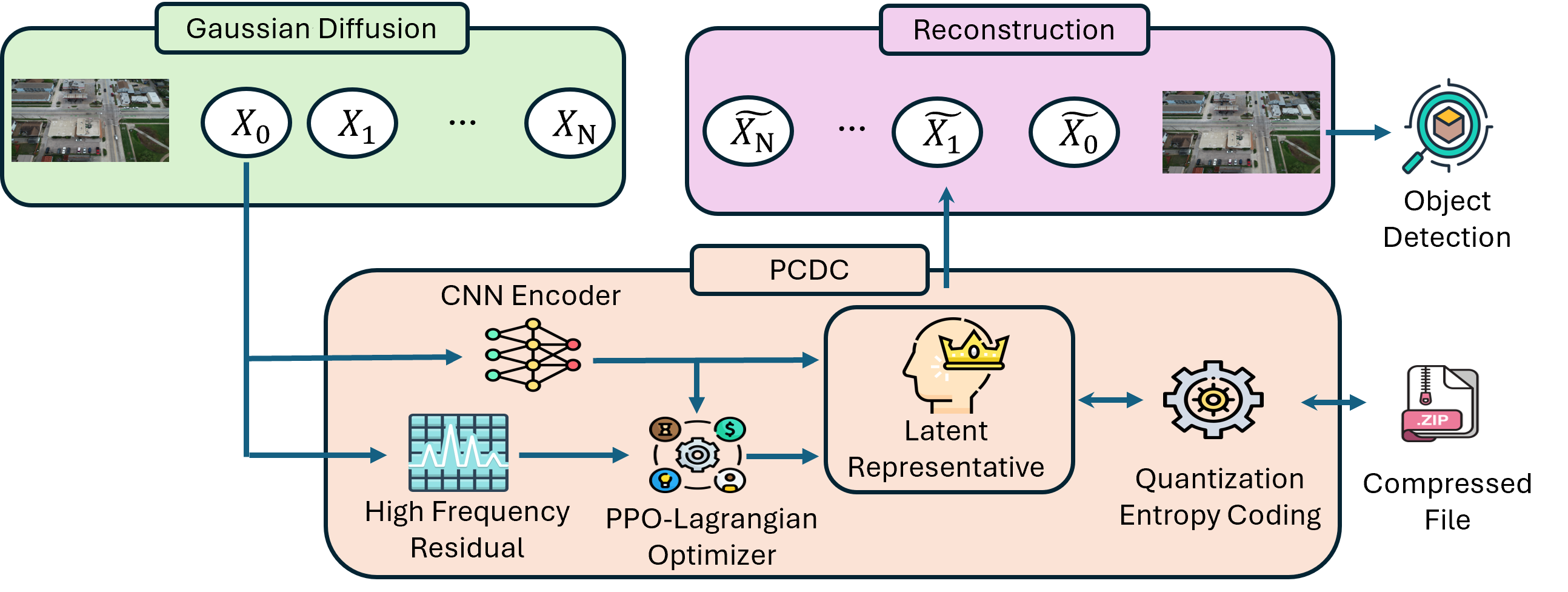}
    \caption{System model of the proposed diffusion-based image compression framework with PPO-based bitrate allocation.}
    \label{fig:system model}
\end{figure*}
\label{sec:model}
\subsection{Conditional Diffusion Based Compression Framework}
\label{sec:cdc_framework}

The proposed framework follows the context-dependent compression paradigm~\cite{cdc}, where the image is first encoded into a latent representation and subsequently reconstructed using a conditional diffusion model. The latent representation captures the global structural and semantic information of the image, while the diffusion model reconstructs the image through iterative denoising conditioned on this latent description. Given the input image \( x_0 \in \mathbb{R}^{H \times W \times 3} \), a CNN encoder extracts a latent representation that summarizes the image content as $\mathbf{z} = \mathrm{Enc}(x_0)$, where \( \mathrm{Enc}(\cdot) \) denotes the encoder network. The latent variable \( \mathbf{z} \) captures the essential spatial structures and contextual information required for image reconstruction while reducing redundancy in the input image.

To enable efficient transmission, the latent representation is quantized using element-wise rounding $\hat{\mathbf{z}} = \mathrm{round}(\mathbf{z})$. The quantized latent \( \hat{\mathbf{z}} \) is then entropy-coded under a learned probability model \( p(\hat{\mathbf{z}}) \). The expected bitrate required to encode the latent representation is therefore given by
\begin{equation}
R_{\mathrm{base}} = \mathbb{E}\left[-\log_2 p(\hat{\mathbf{z}})\right].
\end{equation}
The compressed bitstream corresponding to \( \hat{\mathbf{z}} \) forms the compact representation of the input image and serves as the conditioning signal for the subsequent reconstruction process.

Unlike traditional neural compression systems that employ deterministic decoders, the proposed framework reconstructs the image using a conditional diffusion model. Diffusion models generate samples by reversing a gradual noising process through a sequence of denoising steps. The forward diffusion process progressively perturbs the clean image by adding Gaussian noise as $x_n = \sqrt{\alpha_n}\, x_0 + \sqrt{1 - \alpha_n}\, \epsilon$, where \( \epsilon \sim \mathcal{N}(0, I) \) and \( \{\alpha_n\}_{n=1}^{N} \) defines a predefined variance schedule. During reconstruction, the reverse diffusion process iteratively removes noise to recover the image. The reverse transition is conditioned on the transmitted latent representation \( \hat{\mathbf{z}} \). In practice, the diffusion model is implemented using a U-Net architecture where the latent representation is injected as a conditioning signal to guide the reconstruction process. 

\subsection{PPO-based Bitrate Allocation}
\label{subsec:test_time_drl}

We further propose a constrained deep reinforcement learning strategy to adapt block-wise bit allocation, while satisfying a target compression-ratio constraint. Unlike~\cite{ghouse2023residual}, our encoder does \emph{not} run the diffusion decoder and thus cannot access reconstruction errors. Instead, the policy relies only on available signals: (i) a high-frequency residual map extracted from the input image, and (ii) block-level CNN encoder features. The learned policy outputs block-wise allocation decisions that control local detail preservation during quantization and entropy coding under a global bitrate budget. 
% Policy parameters are updated online at test time using a PPO--Lagrangian algorithm, where decoder-side reconstruction quality provides the learning feedback but does not alter the forward compression pipeline.

First, the encoder computes a high-frequency residual map as $h=\mathcal{H}(x_0)$, where $\mathcal{H}(\cdot)$ is a fixed high-pass filter. We partition the image into $B$ non-overlapping blocks $\{\mathcal{B}_b\}_{b=1}^{B}$. For each block $b$, the policy selects an allocation action $\alpha_b$ controlling local detail preservation. For stability and codec compatibility, we adopt a discrete action space
\begin{equation}
\beta_b \in \mathcal{A}=\{a_1,\ldots,a_K\}, \qquad \boldsymbol{\beta}=\{\beta_b\}_{b=1}^{B}.
\end{equation}
Since the state is constructed from encoder-side observables. At step $b$, we define
\begin{equation}
s_b=\Big[\ \Phi_h\!\big(h|_{\mathcal{B}_b}\big)\ ;\ \Phi_\mathbf{z}\!\big(\mathbf{z}|_{\mathcal{B}_b}\big)\ ;\ \rho_b\ ;\ p_b\ \Big],
\label{eq:state_def}
\end{equation}
where $\Phi_h(\cdot)$ denotes block statistics of the high-frequency residual, $\Phi_\mathbf{z}(\cdot)$ is a pooled CNN embedding for the block, $\rho_b$ is the normalized remaining bitrate budget, and $p_b$ includes block coordinates. We further apply action masking based on the remaining budget to prevent infeasible allocations.

Next, let $R_b(\alpha_b)$ denote the incremental bit cost induced by selecting action $\alpha_b$ for block $b$. The total bitrate is
\begin{equation}
R_{\mathrm{tot}}(\boldsymbol{\alpha}) = R_{\mathrm{base}} + \sum_{b=1}^{B} R_b(\beta_b).
\label{eq:rate_total}
\end{equation}
The target compression-ratio requirement is enforced by a hard bitrate constraint $R_{\mathrm{tot}}(\boldsymbol{\alpha}) \le R_{\max}$.
The diffusion decoder produces a reconstruction $\tilde{x}$. Although the decoder is not invoked in the forward encoder pipeline, it is used to compute learning feedback. We define a utility combining fidelity and perceptual metrics:
\begin{align}
\label{eq:utility}
U(x_0,\tilde{x}_0)= -\lambda_{p}D(x_0,\tilde{x}_0) -\lambda_{s}\big(1-\mathrm{SSIM}(x_0,\tilde{x}_0)\big) \\\notag-\lambda_{l}\mathrm{LPIPS}(x_0,\tilde{x}_0) -\lambda_{d}\mathrm{DISTS}(x_0,\tilde{x}_0),
\end{align}
where $D(\cdot,\cdot)$ is the mean squared error, $\mathrm{SSIM}(\cdot,\cdot)$ denotes the structural similarity index measure, $\mathrm{LPIPS}(\cdot,\cdot)$ is the learned perceptual image patch similarity, and $\mathrm{DISTS}(\cdot,\cdot)$ denotes the deep image structure and texture similarity. To satisfy the bitrate constraint, we optimize a Lagrangian reward:
\begin{equation}
\label{eq:lagrangian_reward}
r = U(x_0,\tilde{x}_0) - \eta\big(R_{\mathrm{tot}}(\boldsymbol{\beta})-R_{\max}\big),
\end{equation}
with dual variable $\eta\ge 0$. The constrained objective is
\begin{equation}
\max_{\pi_\theta}\ \mathbb{E}_{\pi_\theta}\big[U(x_0,\tilde{x}_0)\big]
\quad \text{s.t.}\quad
\mathbb{E}_{\pi_\theta}\big[R_{\mathrm{tot}}(\boldsymbol{\beta})\big]\le R_{\max}.
\label{eq:constrained_objective}
\end{equation}
We solve it by alternating policy optimization and dual ascent:
\begin{equation}
\label{eq:dual_update}
\eta \leftarrow \Big[\eta + \rho\big(R_{\mathrm{tot}}(\boldsymbol{\beta})-R_{\max}\big)\Big]_+,
\end{equation}
where $\rho$ is a stepsize and $[\cdot]_+=\max(\cdot,0)$. This is well suited to the non-monotonic behavior observed in practice, where increasing allocation does not improve perceptual quality.

Last, we parameterize the policy by $\pi_\theta(a_b|s_b)$ and learn a value function $V_\psi(s_b)$. Each episode corresponds to $B$ sequential block decisions. After sampling actions $\{\beta_b\}$ and producing the bitstream, we decode and reconstruct $\tilde{x}$ to compute the terminal reward $r$. Let $G_b=r$ denote the return from step $b$. The advantage estimate is
\begin{equation}
\label{eq:advantage}
A_b = G_b - V_\psi(s_b).
\end{equation}
We define the likelihood ratio $\varrho_b(\theta)=\frac{\pi_\theta(a_b|s_b)}{\pi_{\theta_{\mathrm{old}}}(a_b|s_b)}$.
The PPO clipped surrogate objective is
\begin{align}
\mathcal{L}_{\mathrm{PPO}}(\theta)=\mathbb{E}\Big[\min\big(\varrho_b(\theta)A_b,&\ \mathrm{clip}(\varrho_b(\theta),1-\gamma,1+\gamma)A_b\big)\Big]\\\notag
&+\kappa\,\mathbb{E}\big[H(\pi_\theta(\cdot|s_b))\big],
\label{eq:ppo_objective}
\end{align}
where $\gamma$ is the clipping parameter, $H(\cdot)$ denotes entropy, and $\kappa$ controls exploration. 
% The critic is trained by minimizing
% \begin{equation}
% \mathcal{L}_{V}(\psi)=\mathbb{E}\big[(V_\psi(s_b)-G_b)^2\big].
% \label{eq:value_loss}
% \end{equation}
At test time, we perform a small number of PPO updates per image, followed by the dual update in \eqref{eq:dual_update}. This online adaptation allows the policy to adjust its allocation strategy to the drone images while maintaining the compression constraint through the Lagrangian mechanism. Overall, we outline PPO-Based bitrate allocation in Algorithm~\ref{alg:ppo_lagrangian_short}.

\begin{algorithm}[tbh]
\caption{PPO-Based Bitrate Allocation}
\label{alg:ppo_lagrangian_short}
\begin{algorithmic}[1]
\Require Image $x_0$; HF extractor $\mathcal{H}$; blocks $\{\mathcal{B}_b\}_{b=1}^{B}$; actions $\mathcal{A}$; bitrate constraint $R_{\max}$; policy $\pi_\theta$; value $V_\psi$; dual $\eta$.
\For{each test image $x_0$}:
\State Compute $h\!\leftarrow\!\mathcal{H}(x)$; initialize $\rho_b$.
\For{$b=1$ to $B$}:
\State Form $s_b=[\Phi_h(h|_{\mathcal{B}_b});\Phi_\mathbf{z}(\mathbf{z}|_{\mathcal{B}_b});\rho_b;p_b]$.
\State Sample $\beta_b\sim\pi_\theta(\cdot|s_b)$; update $\rho_b$.
\EndFor
\State Encode to obtain $R_{\mathrm{tot}}$; decode to obtain $\tilde{x}_0$.
\State Compute utility $U$ via \eqref{eq:utility}.
\State Compute reward $r$ by \eqref{eq:lagrangian_reward} and advantages $A_b$ by \eqref{eq:advantage}.
\State Update $\pi_\theta$ with $(s_b,\beta_b,A_b)$; update $\eta$ via \eqref{eq:dual_update}.
\EndFor
\end{algorithmic}
\end{algorithm}

% \subsection{Diffusion Model Network}
% \subsection{High-frequency Information Detection Network}
\section{Numerical Results}
\label{sec:results}
\subsection{Experiments Setup}
\begin{figure}[tbh]
    \centering
    \includegraphics[width=\linewidth]{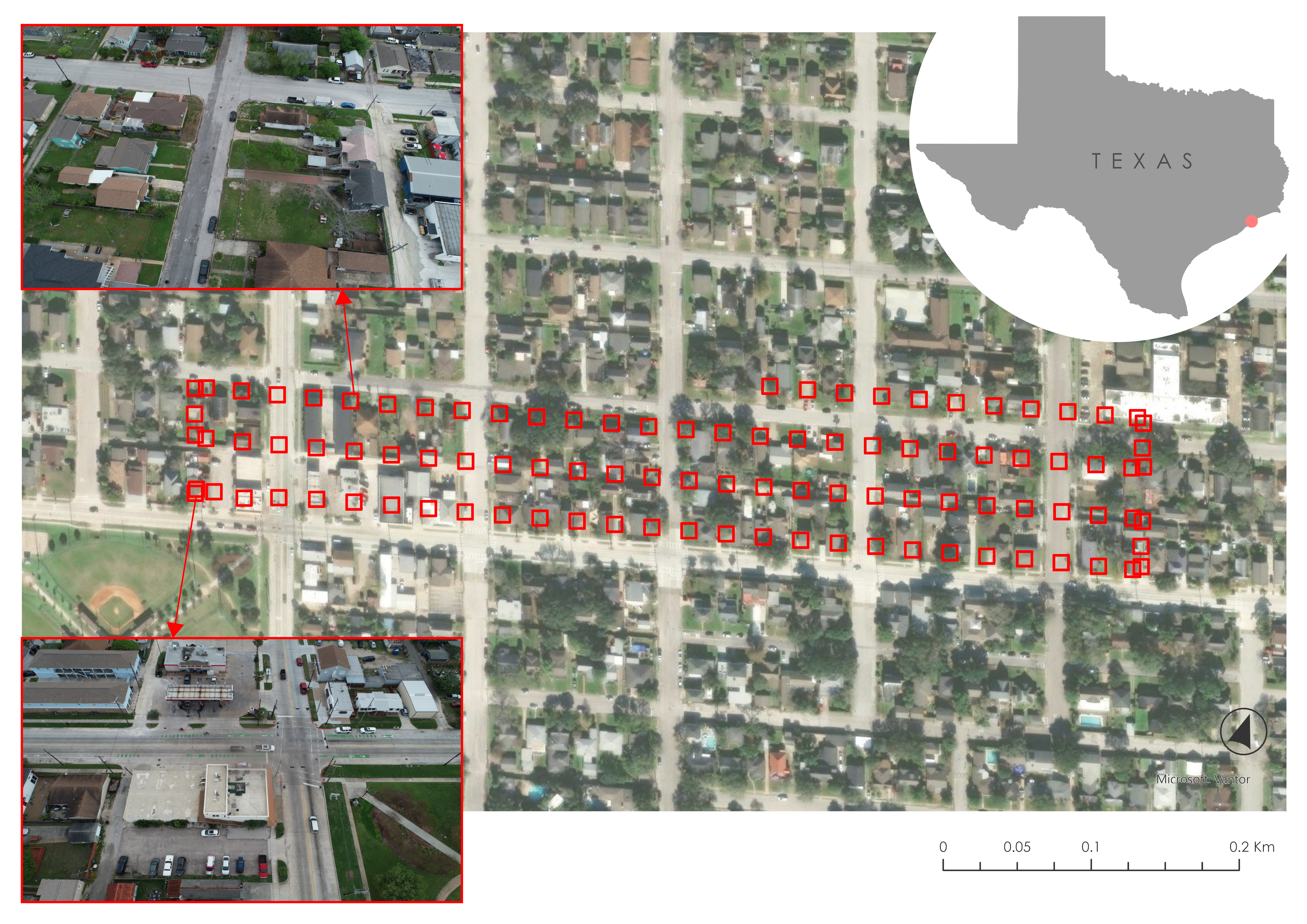}
    \caption{Data collection points and sample images in Galveston, Texas.}
    \label{fig:dataset}
\end{figure}

We evaluate the proposed method on both public DIV2K dataset and our self-collected drone image dataset.
% \footnote{\href{https://huggingface.co/datasets/yuminghan12123/Low-Altitude-Drone-Remote-Sensing-Dataset}{High-Resolution-Low-Altitude-Drone-Remote-Sensing-Dataset}}.at~\cite{our_dataset}
We make this dataset publicly available at~\cite{our_dataset} .
The drone image dataset contains 100 RGB images captured at an altitude of around 47.01~m over urban areas in Galveston, Texas, with a high-resolution of $5472 \times 3648$. As shown in Fig.~\ref{fig:dataset}, it includes nadir views of buildings, roads, parking areas, and vehicles, covering scene characteristics that are not well represented in existing open-source datasets~\cite{zhu2021detection,esri_drone_dataset}.
Each image is partitioned into $16\times16$ blocks. The PPO controller uses a discrete action space with $K=5$ allocation levels to control the preservation strength of each block under bitrate constraint $R_{\max}$, corresponding to a compression ratio of $0.2$. The policy is trained with clipping parameter $\gamma=0.2$, entropy weight $\kappa=0.01$, actor learning rate $3\times10^{-4}$, and critic learning rate $10^{-3}$. The Lagrange multiplier is updated with step size $\rho=10^{-3}$. The reward is weighted by $\lambda_p=1.0$, $\lambda_s=0.5$, $\lambda_l=0.2$, and $\lambda_d=0.2$, respectively.

\subsection{Comparison Results}

\begin{table}[tbh]
\centering
\caption{Compression ratio comparison on DIV2K and drone image datasets.}
\label{tab:compression_ratio}
\begin{tabular}{lcc}
\toprule
\textbf{Model} & \textbf{DIV2K} & \textbf{Drone Image Dataset} \\
\midrule
BPG   & 10.8 & 11.6 \\
HiFiC & 15.9 & 16.8 \\
CDC   & 16.4 & 17.2 \\
PCDC (Ours) & \textbf{19.3} & \textbf{21.2} \\
\bottomrule
\end{tabular}
\end{table}

\begin{figure*}[tbh]
    \centering
    \includegraphics[width=\linewidth]{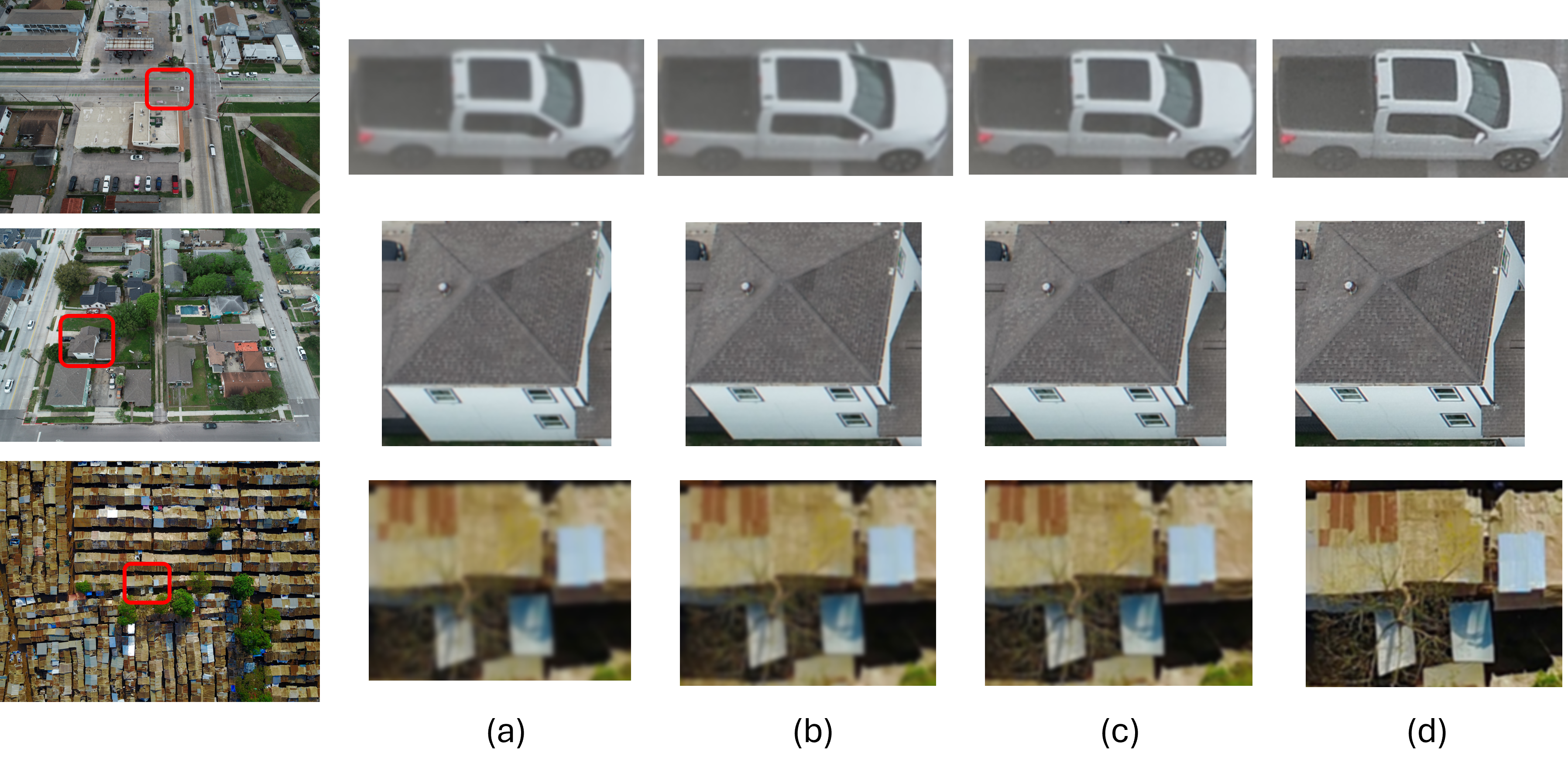}
    \caption{Visual comparison of reconstruction images by different methods. The first two rows are from the proposed drone image dataset and the last row is from DIV2K. (a) BPG, (b) HiFiC, (c) CDC, and (d) the proposed PCDC. The proposed method reconstructs sharper edges and clearer structural details.}
    \label{fig:visual}
\end{figure*}
\begin{figure*}[tbh]
    \centering
    \includegraphics[width=\linewidth]{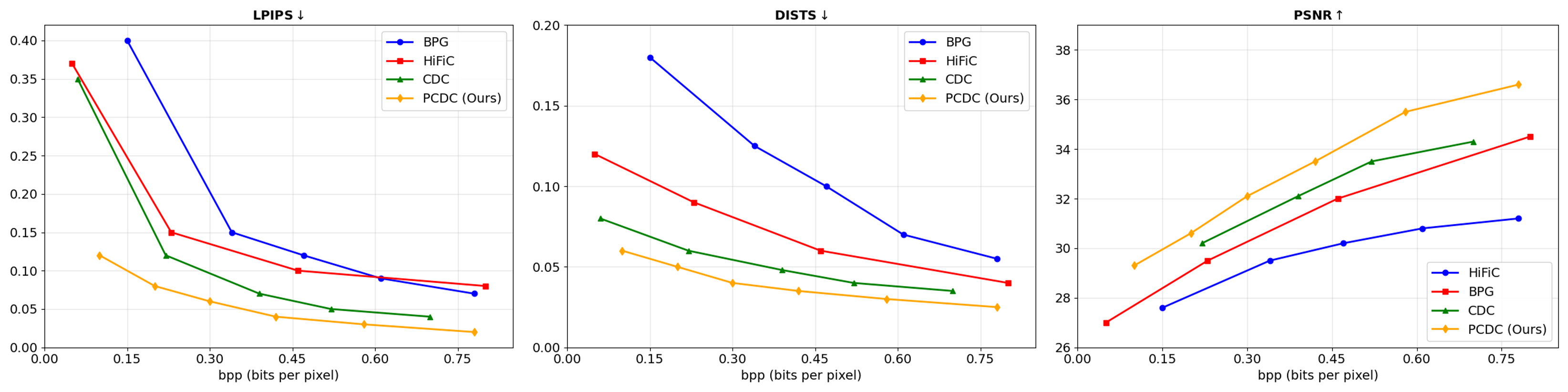}
    \caption{Rate--distortion comparison of BPG, HiFiC, CDC, and the proposed PCDCin terms of LPIPS, DISTS, and PSNR across different bitrates (bpp).}
    \label{fig:psnr_tog}
\end{figure*}

We compare our method to BPG~\cite{BellardBPG2018}, HiFiC~\cite{hific} and CDC~\cite{cdc}.
First, Table~\ref{tab:compression_ratio} shows that the PCDC achieves the highest compression ratio on both datasets. In comparison, BPG yields the lowest compression efficiency, while the learned compression methods HiFiC and CDC provide moderate improvements but remain close to each other. This trend indicates that diffusion-based compression frameworks already offer advantages over traditional codecs, and the proposed enhancements in PCDC further improve coding efficiency. Moreover, all methods achieve higher compression ratios on the drone image dataset, suggesting that aerial imagery with larger homogeneous regions and repetitive structures is  more compressible than the diverse natural scenes. Fig.~\ref{fig:visual} further provides a visual comparison on reconstruction images.

Next, Fig~\ref{fig:psnr_tog} compares the rate–distortion performance of different methods in terms of LPIPS, DISTS, and PSNR across varying bitrates. The proposed PCDC consistently achieves the best performance among all methods. 
% PCDC yields the lowest LPIPS and DISTS values, indicating superior perceptual quality and structural similarity, while achieving the highest PSNR across bitrate levels. 
These results demonstrate that the proposed framework effectively improves both perceptual fidelity and reconstruction accuracy. Moreover, since the PPO-based bitrate allocation is introduced as an add-on to the conditional diffusion compression model, the performance gap between CDC and PCDC can be interpreted as an ablation result. The consistent improvement of PCDC over CDC demonstrates that the proposed bitrate allocation strategy enhances coding efficiency and reconstruction quality beyond the conditional diffusion framework alone.

\subsection{Downstream Object Detection Performances}
For further demonstrating the effectiveness of the proposed model, we perform an object detection downstream task using both pretrained YOLO models (YOLO-pre) and fine-tuned YOLO models (YOLO-ft). We fine-tune the model using an independent drone image dataset separate from the testing set, which includes building labels. Table~\ref{tab:detection_results} reports the detection performance on the original and reconstructed images in terms of the building ratio (B-R), vehicle ratio (V-R), confidence score (Conf.), and intersection over union (IoU), where B and V denote building and vehicle, respectively.
\begin{table}[tbh]
\centering
\caption{Detection results on original and reconstructed images.}
\label{tab:detection_results}
\setlength{\tabcolsep}{4pt}
\begin{tabular}{llcccc}
\toprule
\textbf{Model} & \textbf{Data} & \textbf{B-R} & \textbf{V-R} & \textbf{Conf.} & \textbf{IoU} \\
\midrule
\multirow{2}{*}{YOLO11s-pre}
& Ori.  & --   & 0.99 & V: 0.825 & V: 0.97 \\
& Rec.  & --   & 0.98 & V: 0.828 & V: 0.97 \\
\midrule
\multirow{2}{*}{YOLO11s-ft}
& Ori.  & 1.02 & 0.96 & B: 0.789, V: 0.743 & B: 0.95, V: 0.96 \\
& Rec.  & 1.00 & 0.95 & B: 0.789, V: 0.740 & B: 0.95, V: 0.96 \\
\midrule
\multirow{2}{*}{YOLO12s-pre}
& Ori.  & --   & 0.96 & V: 0.840 & V: 0.97 \\
& Rec.  & --   & 0.95 & V: 0.840 & V: 0.97 \\
\midrule
\multirow{2}{*}{YOLO12s-ft}
& Ori.  & 0.98 & 0.89 & B: 0.753, V: 0.719 & B: 0.96, V: 0.96 \\
& Rec.  & 0.96 & 0.88 & B: 0.751, V: 0.718 & B: 0.96, V: 0.96 \\
\bottomrule
\end{tabular}
% \vspace{0.3em}
% \footnotesize{B: building; V: vehicle; B-R: average building ratio; V-R: average vehicle ratio; Conf.: average detection confidence score; IoU: intersection over union; YOLO-pre: pretrained YOLO model; YOLO-ft: fine-tuned YOLO model.}
\end{table}
\begin{figure}[tbh]
    \centering
    \includegraphics[width=\linewidth]{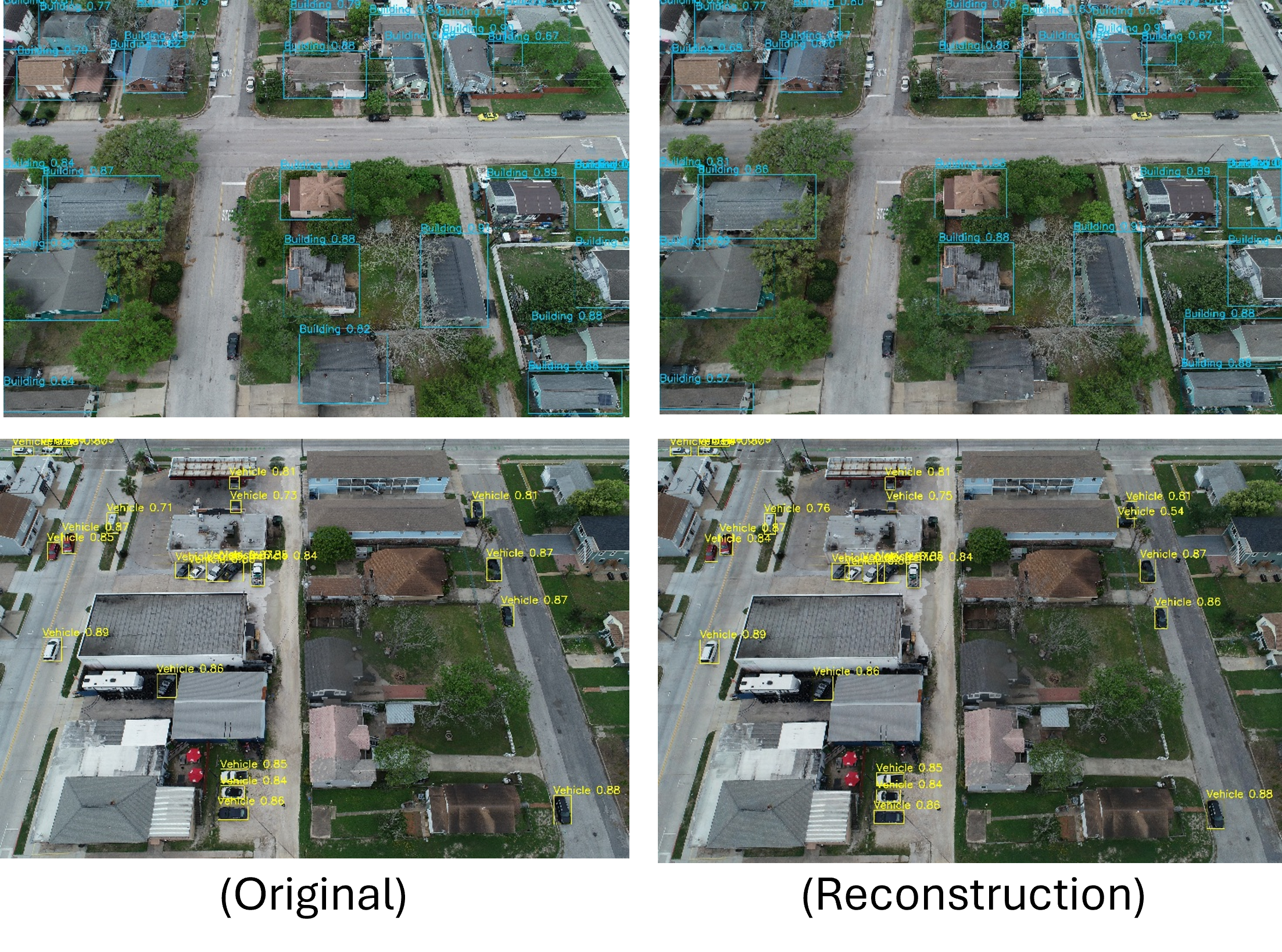}
    \caption{Visual comparisons with corresponding confidence scores for building (first row) and vehicle (second row) detections.}
    \label{fig:reconstuction}
\end{figure}
 Fig.~\ref{fig:reconstuction} further provides a visual comparison with confidence score on vehicles and buildings.
% The results show that the detection statistics on reconstructed images remain very close to those obtained from the original images across all models. In particular, the average vehicle and building ratios exhibit only marginal changes, while the confidence scores and IoU values remain nearly unchanged. 
These negligible average differences of around 0.02 in Table~\ref{tab:detection_results} indicate that the proposed compression framework preserves important semantic structures and object-level features, allowing downstream detection tasks to maintain comparable performance after reconstruction.

\section{Conclusion}
\label{sec:conclusion}
In this paper, we proposed PCDC, a conditional diffusion-based framework for RS image compression that preserves reconstruction quality and downstream task utility. With a novel PPO-based bitrate allocation strategy, PCDC achieves high compression ratios while maintaining strong perceptual performance. In addition, we release a high-resolution drone image dataset and demonstrate through downstream object detection experiments that the reconstructed images preserve task-relevant information with negligible performance loss. These results suggest that PCDC can support efficient storage planning and long-term data management by reducing the storage costs associated with large-scale drone imagery archives. Future work will extend the framework to more remote sensing tasks and investigate more efficient diffusion sampling for practical deployment.

% \section{Acknowledgement}
% The authors appreciate the Disaster Data Research Center (DDRC) of Texas A\&M University at Galveston for providing access to the drone imagery used in this study. The DDRC facilitates the collection and sharing of disaster-related datasets that support research in hazard analysis and resilience. Additional information on DDRC is available at https://idrt.tamug.edu/ddrc/

\bibliographystyle{IEEEtran}
\bibliography{references}

% \newpage

 

% \end{IEEEbiography}



% \vfill

\end{document}